%% file: root.tex
\documentclass[letterpaper, 10 pt, conference]{ieeeconf}
\IEEEoverridecommandlockouts                             
\include{includes/0-preamble}

\title{\LARGE \bf
Object Rearrangement Using Learned Implicit Collision Functions
}

\author{Michael Danielczuk*$^{1,2}$, Arsalan Mousavian*$^{1}$, Clemens Eppner$^{1}$, Dieter Fox$^{1,3}$%
\thanks{$^{1}$NVIDIA, USA. $^{2}$UC Berkeley. $^{3}$University of Washington, Paul G.~Allen School of Computer Science \& Engineering, Seattle, WA, USA.
        {\tt\small mdanielczuk@berkeley.edu, amousavian@nvidia.com, ceppner@nvidia.com, dieterf@nvidia.com}}%
}

\begin{document}
\maketitle
\thispagestyle{empty}
\pagestyle{empty}

\begin{abstract}
Robotic object rearrangement combines the skills of picking and placing objects. When object models are unavailable, typical collision-checking models may be unable to predict collisions in partial point clouds with occlusions, making generation of collision-free grasping or placement trajectories challenging. We propose a learned collision model that accepts scene and query object point clouds and predicts collisions for 6DOF object poses within the scene. We train the model on a synthetic set of 1~million scene/object point cloud pairs and 2~billion collision queries. We leverage the learned collision model as part of a model predictive path integral (MPPI) policy in a tabletop rearrangement task and show that the policy can plan collision-free grasps and placements for objects unseen in training in both simulated and physical cluttered scenes with a Franka Panda robot. The learned model outperforms both traditional pipelines and learned ablations by $9.8\%$ in accuracy on a dataset of simulated collision queries and is 75x faster than the best-performing baseline. Videos and supplementary material are available at~\url{https://research.nvidia.com/publication/2021-03_Object-Rearrangement-Using}.
\end{abstract}

\input{includes/1-introduction}
\input{includes/2-related-work}

\input{includes/3-problem-statement}
\input{includes/4-methods}
\input{includes/5-policy}
\input{includes/6-collision-results}
\input{includes/7-policy-results}
\input{includes/8-conclusion}


%%%%%%%%%%%%%%%%%%%%%%%%%%%%%%%%%%%%%%%%%%%%%%%%%%%%%%%%%%%%%%%%%%%%%%%%%%%%%%%%

\renewcommand*{\bibfont}{\footnotesize}
\clearpage
\printbibliography

\end{document}

%% file: includes/0-preamble.tex
\usepackage{hyperref}
\usepackage[backend=biber,
            hyperref=true,
            url=false,
            isbn=false,
            doi=false,
            backref=false,
            style=ieee,
            natbib=true,%compatibility aliases
            mincitenames=1,
            maxcitenames=1,
            citestyle=numeric-comp,
            sorting=nyt,%none
            block=none]{biblatex}
        
\usepackage[dvipsnames]{xcolor}
\usepackage{xspace}
\usepackage{autolabtools}

\usepackage{graphicx}
\graphicspath{ {figures/} }
\usepackage{tabu}
\usepackage{booktabs}

\usepackage{amsmath}
\usepackage{amsfonts}
\usepackage{bm} % for bolding math

\DeclareMathOperator*{\argmin}{arg\,min}

\newcommand{\algname}{SceneCollisionNet\xspace}

\usepackage{etoolbox}
\makeatletter
\patchcmd{\@makecaption}
  {\scshape}
  {}
  {}
  {}
\patchcmd{\@makecaption}
  {\\}
  {.\quad}
  {}
  {}
\patchcmd{\@makecaption}{\par}{\par\justifying}{}{}
\def\fnum@table{Table~\thetable}
\makeatother

%% file: includes/1-introduction.tex
\section{Introduction}

%%% (1) Motivate Problem: Collision checking for rearrangement
Rearranging objects is a fundamental robotic skill~\cite{batra2020rerrangement} with broad impacts in applications ranging from logistics in industrial settings to service robotics at home. The majority of existing approaches rely on known models of the objects and environment to generate collision-free trajectories for grasping, moving, and placing objects in a scene~\cite{king2016rearrangement,huang2019large}. When only point cloud data for the scene and objects are available, these approaches may not correctly reason about occlusions or quickly react to a changing environment. We focus on a core ability that enables rearrangement of unknown objects in cluttered unknown environments: collision checking based on raw sensor measurements.

%%%DDD
Existing techniques for collision checking between objects and scenes are limited in that they either rely on known object models or struggle to reason about occluded areas of a scene~\cite{gilbert1988fast,pan2012fcl,pan2011probabilistic}. Recently, the computer vision community has introduced deep learning techniques with astonishing abilities to represent and reason about fine-grained 3D object geometries~\cite{park2019deepsdf,chabra2020deep,jiang2020local}. Unfortunately, these approaches are not efficient enough to handle the large number of collision queries necessary for efficient trajectory optimization and control in robotics. In this paper, we introduce an approach that overcomes these limitations and provides robust collision checking on point clouds with occlusions at speeds that are beyond model-based collision checkers used in robotics. 

\begin{figure}
    \centering
    \includegraphics[width=\linewidth]{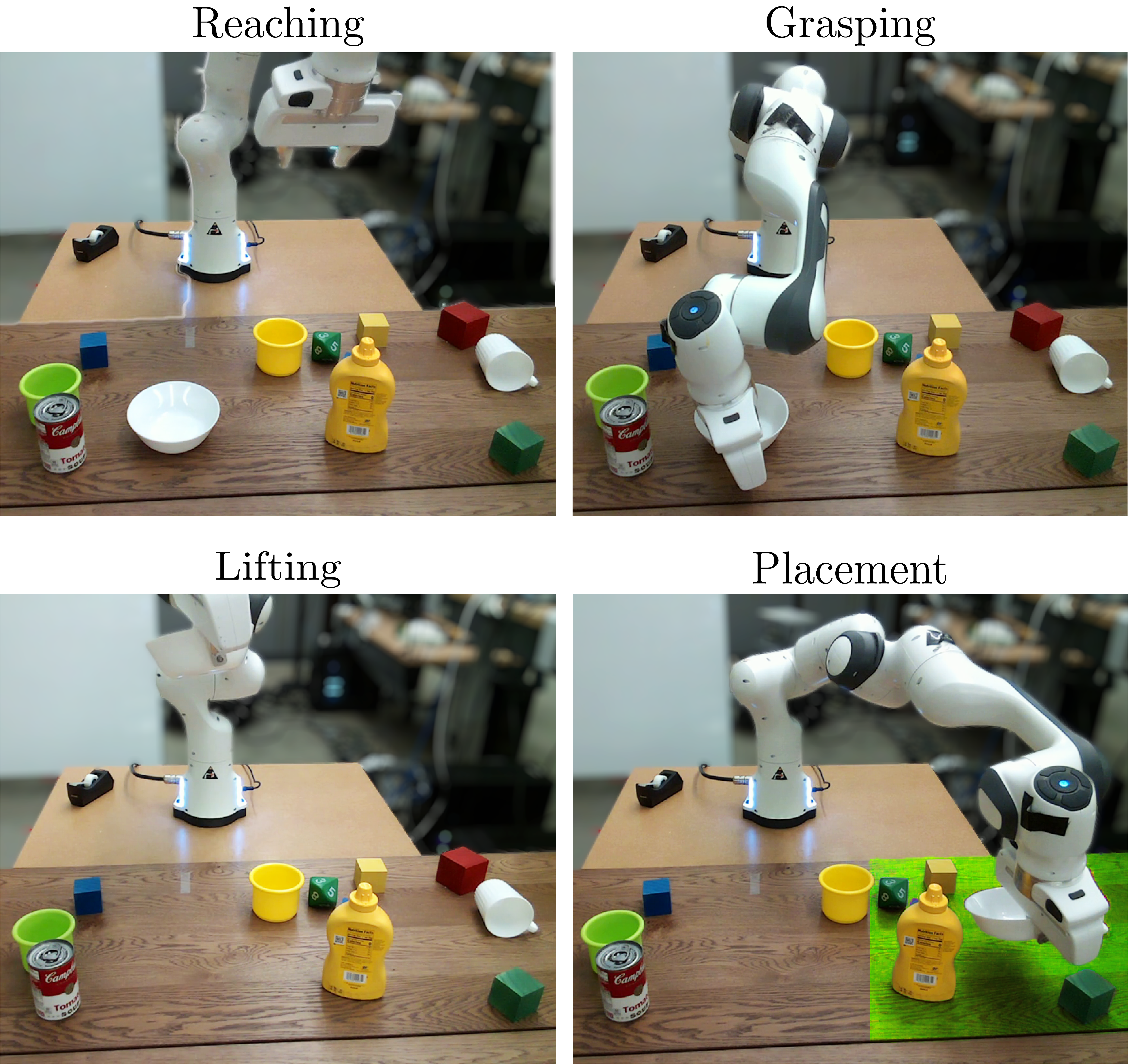}
    \caption{The rearrangement task consists of four subtasks: reaching for, grasping, lifting, and placing the object within a placement zone (overlaid in green). In each subtask, the robot must plan a kinematically feasible, collision-free path with only partial point cloud observations.}
    \label{fig:rearrangement}
    % \vspace{-3.5mm}
\end{figure}

%%% (3) How this work solves the problem
We present a neural network that takes as input raw point clouds of both an object and a scene and a 6DOF pose of the object in the scene and outputs the likelihood that the object collides with the scene. We combine point features with voxel features to construct a scene representation that is both fast and memory efficient. We train the model entirely in simulation with 1 million randomly generated tabletop scenes and show it can generalize to real point cloud data. The resulting model can be used in any existing motion planning framework to generate collision-free motion plans; we demonstrate its capability for object rearrangement via pick-and-place actions based on point cloud measurements in a model predictive control framework with no additional learned parameters.

%%% (4) Contribution summary
This paper makes three contributions:
\begin{enumerate}
\item \algname: a model architecture and training procedure for collision checking between point clouds.
\item A rearrangement policy using \algname in a model predictive path integral controller.
\item Experimental results in simulation and on a real robot platform showing \algname achieves 93\% accuracy on 2 million object-scene queries, taking only 10 $\mu$s per query, 75x faster than baselines.
\end{enumerate}

%% file: includes/2-related-work.tex
\section{Related Work}

\subsection{Robot Collision Detection from Point Clouds}
When mesh models for objects in a scene are known, there exist fast and accurate methods for checking collisions between robot links and the scene or between objects themselves~\cite{pan2012fcl,gilbert1988fast}. However, in scenes containing unknown objects, only partial point cloud data may be available. One approach to collision checking for point cloud data is to expand each point as a sphere with a predefined radius~\cite{hubbard1996approximating}, but the radius may be difficult to determine and may affect the resolution of the collision queries. Similarly, voxel-based approaches are memory-intensive~\cite{cciccek20163d} and can suffer from resolution errors due to discretization. Bounding volume hierarchies that attempt to capture a representation of the shape from the points have also been considered~\cite{figueiredo2012collision,klein2004point}, but again may not capture occluded areas and may not be robust to noise in the point cloud. \citet{pan2011probabilistic} cast the problem as a binary classification problem, use an SVM to learn a boundary surface between the point clouds, and determine collision probability based on the probability of points crossing the boundary. In contrast, we encode the scene into a set of latent voxel vectors instead of checking collisions between raw point clouds and show an ability to reason about partially observable areas in real time.

\subsection{Point Cloud Surface Representations}
Another approach to point cloud collision detection is to derive a representation of the underlying surface and check collisions against that representation. Adaptive meshes~\cite{terzopoulos1991sampling} or alpha shapes~\cite{bajaj1995automatic,edelsbrunner1994three} convert an unstructured array of 3D points into a triangular or tetrahedral mesh. \citet{berger2017survey} provide an excellent survey of surface reconstruction methods. Data-driven approaches also reconstruct underlying representations from point clouds or depth images~\cite{song2017semantic,groueix2018papier,dai2019scan2mesh}. They typically encode points using either point~\cite{qi2017pointnet,qi2017pointnet++} or voxel~\cite{wu20153d,zhou2017voxelnet} representations, or a combination of the two~\cite{liu2019point}. Several recent approaches use fully-connected neural networks to encode an implicit representation of the surface as a function in 3D space~\cite{park2019deepsdf,chen2019learning,mescheder2019occupancy}, showing an ability to reconstruct objects with fine geometries. \citet{van2020learning} similarly reconstruct objects from partial point clouds without optimizing for a latent vector at run time. \citet{jiang2020local} and \citet{chabra2020deep} further encode the surface into many latent vectors across discrete voxels as opposed to a single latent vector for the shape for better scene-level performance. We similarly discretize space into voxels and encode the points in each voxel into a latent vector, but optimize end-to-end for collision queries instead of reconstructing the underlying object geometry.

\subsection{Accelerating Collision Detection}
As collision checking is considered one of the bottlenecks in motion planning, several methods accelerate it using previously calculated collision results~\cite{pan2016fast,kumar2019learning}. \citet{pan2012gpu} developed a GPU implementation of bounding volume test tree traversal that dramatically increases the speed of generating collision-free motion plans for a PR2 robot. Fastron~\cite{das2020learning} and ClearanceNet~\cite{kew2019neural} generate $\mathcal{C}$-space models for collision checking. ClearanceNet also batches collision checks and does not need retraining when objects move. \citet{tran2020predicting} use a contractive autoencoder and multi-layer perceptron to predict collisions in latent space between a robot and axis-aligned boxes. However, each of these methods assumes knowledge of object geometry whereas our method operates directly on point cloud data.

\subsection{Robotic Object Rearrangement}
There has been considerable work on planning for object rearrangement in tabletop scenarios~\cite{lee2019efficient,shome2018fast,king2016rearrangement,huang2019large}; however, these approaches typically rely on known models of both the environment and the objects in the scene for finding collision-free grasping and placement motions. Recently, there have been advances in 6-DOF grasping~\cite{mousavian20196,murali20206} and closed-loop grasping~\cite{song2020grasping,morrison2020learning};~\citet{murali20206} learn collisions between grippers and cluttered scenes centered around a target object, but their method does not easily lend itself to broader motion planning frameworks. \citet{gualtieri2018learning} learn pick and place actions for block, mug, and bottle objects, but use top-down point clouds and do not account for workspace dynamics. The Amazon Picking Challenge has also focused development of pick and place systems~\cite{zeng2018robotic}. In contrast, we integrate a learned point cloud collision checker into an existing motion planning framework for both grasping and placement actions. \citet{zeng2020transporter}, \citet{yuan2018rearrangement}, and \citet{haustein2019learning} learn policies to pick and place or rearrange objects via grasping or pushing directly from input images similar to our approach~\cite{zakka2020form2fit,stevvsic2020learning}. However, they consider planar tasks that do not require the robot to reason about 3D collisions with other objects.

%% file: includes/3-problem-statement.tex
\section{Problem Statement}
We consider a problem setting where a robot with a parallel-jaw gripper iteratively grasps and places objects on a tabletop to rearrange them. The objective is to rearrange the objects as quickly as possible while reacting to changes in the environment. Observations of the scene are given by a single depth sensor with known camera intrinsics pointing toward the table and robot at an oblique angle.

\begin{figure*}[ht!]
    \vspace{1.5mm}
    \centering
    \includegraphics[width=\linewidth]{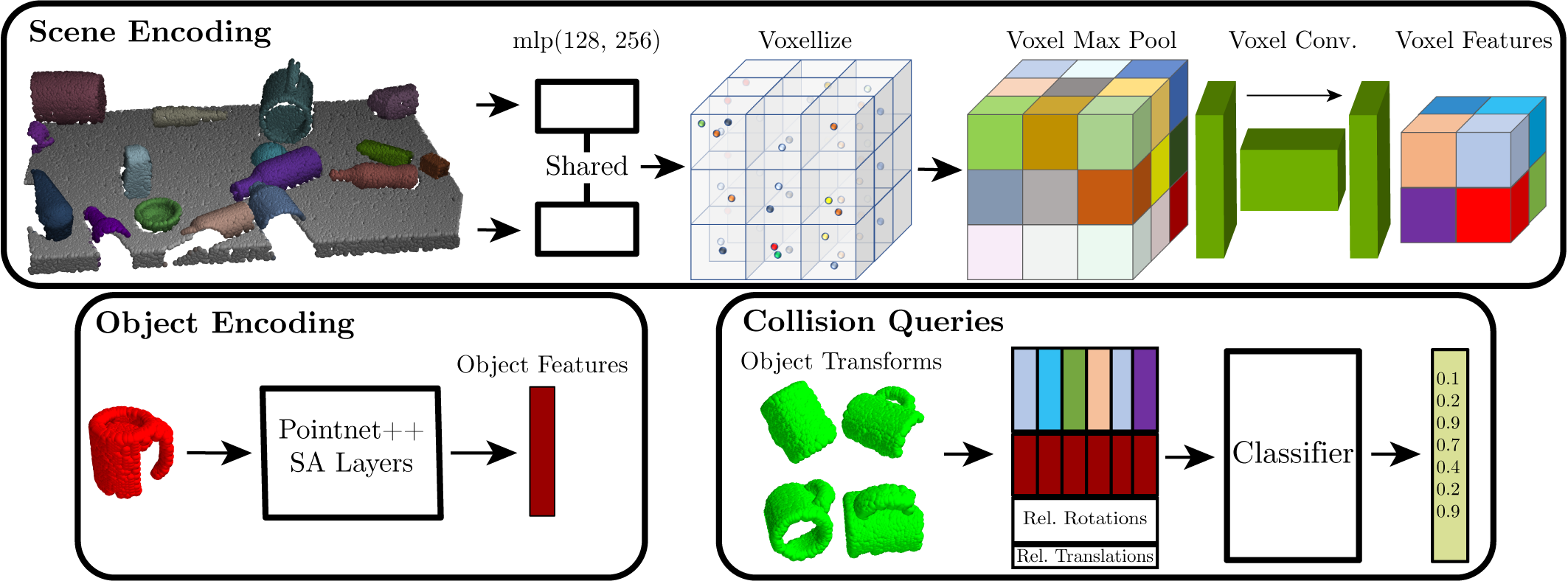}
    \caption{Network architecture for \algname, which predicts collisions between a scene point cloud and an object point cloud given a relative 6DOF pose of the object within the scene. Scene points are encoded by voxelizing, featurizing, and convolving the max-pooled voxels. Object points are encoded using Pointnet++~\cite{qi2017pointnet++} layers. Collision queries are created by feeding the concatenated voxel features, object features, and the relative object transform into a small classifier that predicts the likelihood of collision. Object transforms are specified relative to the voxel frame such that collision queries across different voxels can be predicted simultaneously.}
    \label{fig:network-diagram}
    \vspace{-2mm}
\end{figure*}

\subsection{Definitions} \label{subsec:defs}
We define the problem as having:
\begin{itemize}
    \item \textbf{States} $(S)$: A state $\mathbf{s}_k$ at time $k$ consists of a valid robot joint configuration $\mathbf{q}_k$ and a tabletop containing $N$ objects. No prior information is known about the $N$ objects. $S_{k, \text{free}} \subset S$ is the set of collision-free states at time $k$ and $G_k \subset S$ is the set of goal grasp or placement states.
    \item \textbf{Observations} $(O)$: An observation $\mathbf{y}_k \in \mathbb{R}^{n \times 3}$ at timestep $k$ consists of a point cloud with $n$ points from the camera pointing at the scene.
    \item \textbf{Actions} $(A)$: Actions are defined as a change in the joint configuration of the robot $\mathbf{a}_k = \Delta \mathbf{q}_k$.
    \item \textbf{Transitions} $(T)$: The transition model $T(\mathbf{s}_{k+1} \ | \ \mathbf{a}_k, \mathbf{s}_k)$ represents the dynamics of the scene and robot and is executed by Isaac Gym in simulation~\cite{liang2018gpu}. On the physical system, next states are determined by executing the action on a physical robot.
    \item \textbf{Cost Function} $(C)$: The cost of a state $C(\mathbf{s}_k)$ is defined as the minimum L2 distance from the current robot joint configuration to the goal robot joint configuration: $C(\mathbf{s}_k) = \min_{\mathbf{g}_k} \ \|\mathbf{s}_k - \mathbf{g}_k\|_2 \ s.t. \ \mathbf{g}_k \in G_k$.
\end{itemize}

\subsection{Objective}
The rearrangement objective is to find a policy $\pi$ that minimizes the total cost of the states visited during grasping and placement over a finite horizon $H$ subject to kinematic constraints and that all states along the trajectory are collision-free:
\begin{align*}
    \pi^* = \argmin_\pi \ \mathbb{E}_{\mathbf{a}_k \sim \pi(\mathbf{s}_k)} \sum_{k=1}^H C(\mathbf{s}_k) \ s.t. \ \mathbf{s}_k \in S_{k, \text{free}}
\end{align*}

%% file: includes/4-methods.tex
\section{SceneCollisionNet}
To predict collisions between two point clouds, we propose \algname, a deep neural network inspired by recent work in implicit surface representations from point clouds. Similar to~\citet{jiang2020local} and~\citet{chabra2020deep}, we divide space into coarse voxels and use a local representation for each voxel based on the points contained within that voxel. However, our experiments show that for collision queries: (1) explicitly reconstructing the underlying surface within each voxel is unnecessary and (2) scene information must be shared between voxel representations. We also avoid the costly latent vector optimization, enabling real-time collision prediction. 

Our model divides the scene point cloud into coarse voxels (side length of about $10\,cm$) and assigns points to the voxels, normalizing each point within its voxel by subtracting the voxel's center. We pass the points through a shared multi-layer perceptron and max-pool the features of the points per voxel, similar to Pointnet~\cite{qi2017pointnet}. The max-pooled voxel features are passed through 3D convolution layers, similar to~\citet{liu2019point}, incorporating global information from neighboring voxels. The target object point cloud is featurized separately using Pointnet++ set abstraction layers~\cite{qi2017pointnet++}. Collision queries consist of the transform of the object relative to the nearest voxel center, the corresponding voxel features, and the object features. This approach means the scene and object features are generated only once per scene point cloud and a large number of collision queries can be made in a single forward pass through the classifier, which predicts the likelihood of collision for each transformation. Figure~\ref{fig:network-diagram} shows the network architecture.

\subsection{Dataset Generation and Training}
We train \algname entirely using synthetic point clouds. For each scene, we place objects drawn from a dataset of 8828 3D mesh models~\cite{eppner2021icra} in one of their stable poses with a uniformly random rotation applied about the world $z$-axis on a planar surface. Object positions are chosen uniformly at random such that they do not collide with other objects. We draw the number of objects from a uniform distribution between 10 and 20. The camera, which renders a scene point cloud, is aimed at the origin of the scene and its extrinsics are taken from uniform distributions centered at their nominal values.
A query object is also drawn from the dataset of mesh models; this object is placed at the origin in a random stable pose, where a point cloud is rendered using the same camera. We then generate $q$ collision queries by moving the query object along $t$ trajectories through the scene, recording its relative rotation, translation, and ground truth collisions with the scene using the flexible collision library (FCL)~\cite{pan2012fcl}. The trajectory is formed by linearly interpolating between the start and end object poses, which are chosen randomly. Generating one scene/target pair with $q = 2048$ queries over $t = 64$ trajectories takes roughly 2 seconds on an Ubuntu 18.04 machine with an Intel Core i7-7800X 3.50GHz CPU. 
Each epoch of training consists of 1,000~unique scene/object/trajectory inputs and we train each model for 1000~epochs, or a total of 1~million unique inputs and just over 2~billion total collision queries. We adopt a hard negative mining scheme, where we backpropagate the loss only from the 10\% highest loss queries plus 10\% random queries, which increases the true positive rate by 6\% for similar accuracy. Training takes about 9~days on an NVIDIA V100 GPU. We use SGD with learning rate $1e-3$ and momentum $0.9$.

\subsection{Robot Collision Checking} \label{sec:robot-collision-checking}
For robot collision checking, we pre-sample points from the 3D mesh of each link in the robot's kinematic chain and featurize each set of points. This feature set is only generated once for a given robot. The set of link features and link poses (using forward kinematics for a given configuration) are input to \algname with the scene features at run time; collision predictions can then be generated for all links in a single forward pass. The same method can also be used to predict collisions between other known meshes and a partial scene point cloud, showcasing the flexibility of our method.

%% file: includes/5-policy.tex
\section{Object Rearrangement}
Rearrangement of objects is a multi-stage task, so we incorporate a finite state machine into our policy with 5 states: reaching the pre-grasp pose, attempting the grasp, lifting the object, placing the object, and releasing the placed object. We use a model predictive path integral (MPPI) policy for the reaching and placing states and preset actions for reaching from the pre-grasp to final grasp pose, lifting, and releasing the object. We use \algname to find both placement positions and collision-free trajectories for grasping and placing.

\subsection{Grasps and Placements}
\begin{figure}[t!]
    \vspace{1.5mm}
    \centering
    \includegraphics[width=\linewidth]{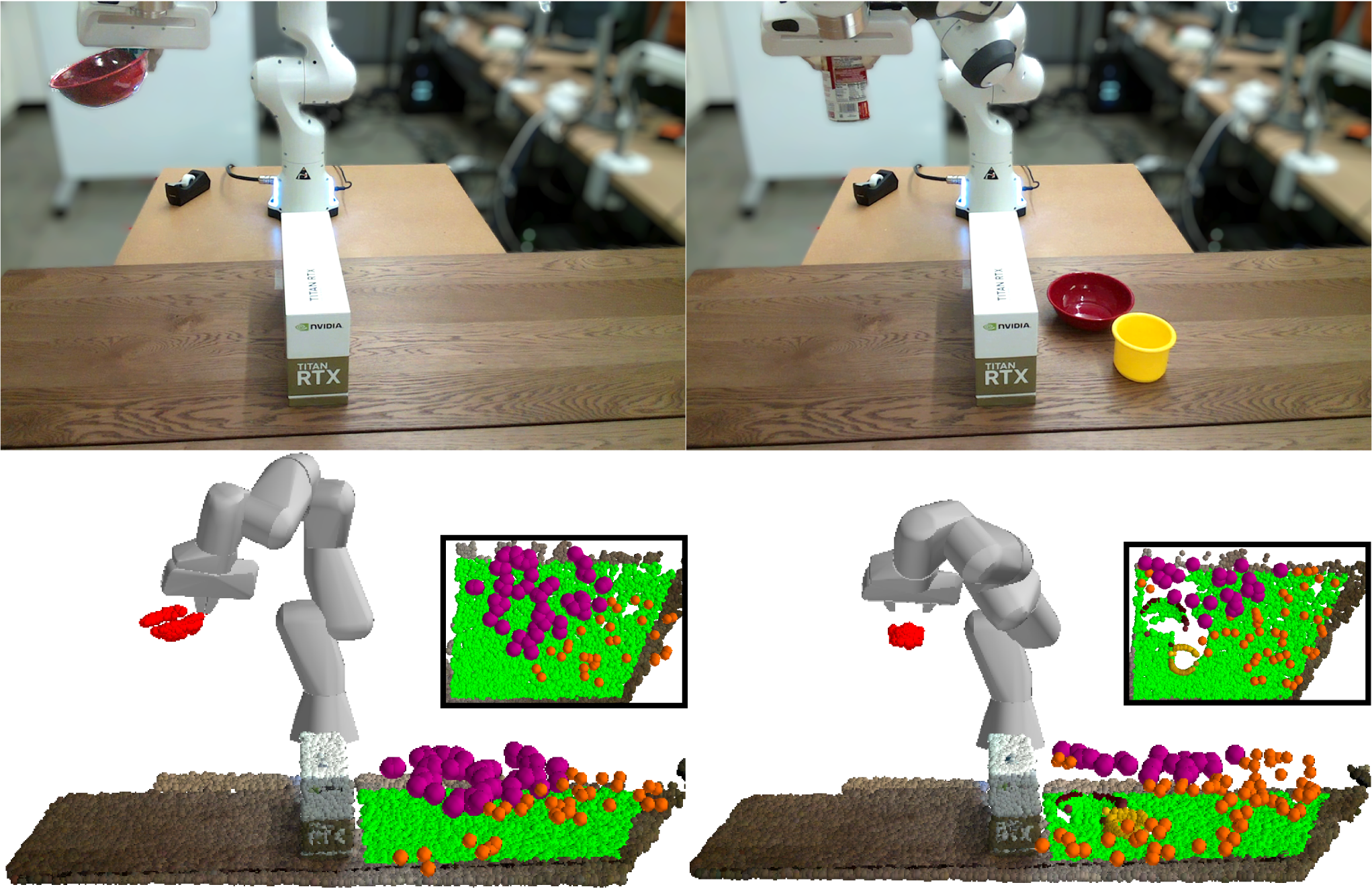}
    \caption{With an uncluttered (left) placement zone (green), collision-free placement goals with inverse kinematics solutions (purple dots) spread across the zone within reach of the robot. When the zone is cluttered (right), \algname predicts placements around the objects. Orange dots represent collision-free placements without IK solutions.}
    \label{fig:placements}
    % \vspace{-1mm}
\end{figure}

We modify Contact-GraspNet~\cite{sundermeyer2021icra} to predict 6DOF grasps on a region of the raw point cloud in cluttered environments and the segmentation from~\citet{xiang2020learning}. We use the Trac IK solver~\cite{beeson2015trac} to convert grasp poses to robot configurations. We accept a point cloud mask that represents an area of the scene where the object should be placed, which by default includes the entire workspace. Points are sampled uniformly at random within the placement zone and sorted by height in the scene; \algname classifies whether the object would be in collision at the given point and the lowest collision-free points are chosen as placement goals. Figure~\ref{fig:placements} shows placement candidates for both empty and cluttered placement zones for the object in hand. Final placement goals (purple) must be both collision-free and have an inverse kinematics solution. Orange points show placement candidates without inverse kinematics solutions; this decoupling allows for the same placements to be used with a different robot.

\subsection{MPPI Policy}
We leverage the parallelism provided by \algname in a model predictive path integral (MPPI) algorithm for object rearrangement in tabletop environments~\cite{williams2017model}. The advantages of MPPI in this setting are: 1) the task can be specified entirely in the joint space and joint constraints can be strictly enforced during trajectories, 2) trajectory costs can easily be specified using distances in joint space, 3) trajectory generation, cost calculation, collision checking, and forward kinematics can be parallelized on a GPU for the real-time capability necessary in closed-loop execution. In contrast, standard motion planning techniques, such as RRT or PRM, may provide guarantees of completeness and optimality, but are by nature sequential, require a nearest neighbor search for connecting nodes, and must be adapted for dynamic environments.

We adapt MPPI such that trajectories are generated by sampling around a linear trajectory between the start and goal joint configurations. Specifically, we create $T$ vectors by perturbing the straight-line trajectory $\mathbf{d}$ with a vector drawn from a normal distribution and renormalizing: $\Tilde{\mathbf{d}}_i = N(\mathbf{d} + \mathcal{N}(\mathbf{0}, \Sigma))$. Trajectories consist of $H$ steps along $\Tilde{\mathbf{d}}_i$; actions are clipped to the robot joint limits at each timestep.

\begin{figure}[t!]
    \vspace{1.5mm}
    \centering
    \includegraphics[width=\linewidth]{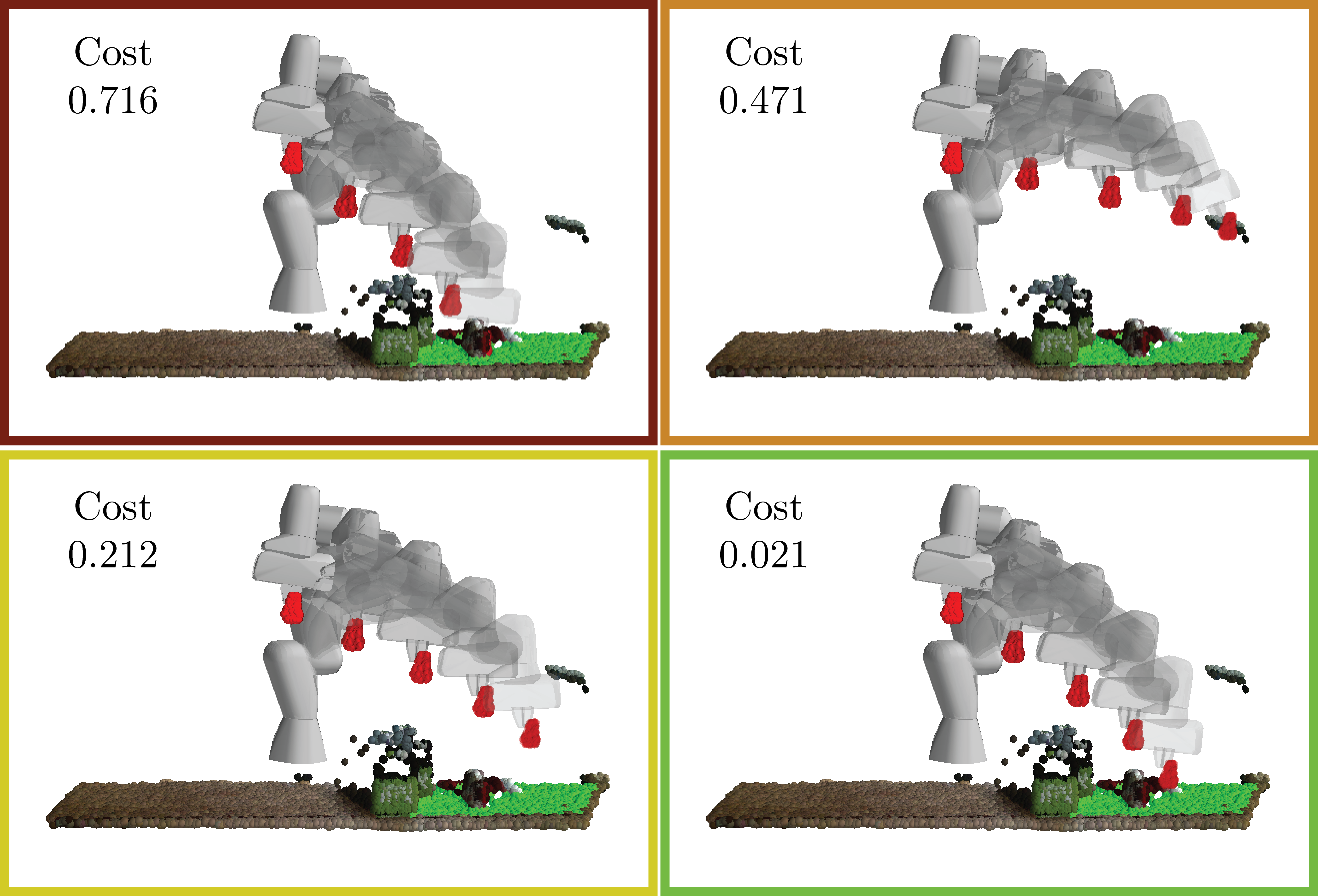}
    \caption{Sampled trajectories from the current robot configuration (solid) to the ending configuration (transparent) with outline colors indicating the cost of each trajectory. The lowest cost trajectory is collision-free and brings the object close to the placement area.}
    \label{fig:rollouts}
\end{figure}

The cost of each trajectory is the cost of its final state as defined in Section~\ref{subsec:defs}. We check both robot-scene collisions as in~\ref{sec:robot-collision-checking} and robot self-collisions using a model that predicts distance to self-colliding configurations~\cite{rakita2018relaxedik} at discrete intervals between each waypoint in each trajectory. Thus, at each policy call, we make $T \times H$ collision checks for each robot link, which can be computed in a single forward pass using \algname. If there is an object in hand, collisions between the object and the scene are also checked at each point in the trajectory. Then, we remove all waypoints after the first colliding waypoint and clip the trajectory to the waypoint with minimum cost. The minimal cost trajectory is executed until the policy is called again. Figure~\ref{fig:rollouts} shows a sampling of four trajectories and their associated costs.

Importantly, the scene points belonging to the robot or to the target object must be removed during placement; if they remain in the scene, they will cause all MPPI trajectories to be in collision. In physical experiments, we combine a learned robot point cloud segmentation model and a particle filter to track the robot points and remove them. We segment the target object before grasping it and remove points within an object-tracking box that is transformed relative to the end-effector. Note that this approach does not account for any in-hand motion of the object, but avoids end effector occlusions after grasping. In simulation, we use ground truth segmentation masks for both the robot and the target object.

%Since the robot moves through the scene, large parts of the scene may be occluded due to the single camera view, making collision predictions in those parts of the scene unreliable. Thus, we aggregate point clouds across timesteps by replacing points that are occluded by the robot with points from previous timesteps. When those areas of the scene are no longer occluded, we again use the current points. Both this aggregation process and the placement process require segmentation of the scene - aggregation requires the robot points be segmented  and placement requires that the object to be placed be segmented so that its point cloud can be queried against the scene representation for collisions. \noteby{AM}{MD: need some merging here}

%% file: includes/6-collision-results.tex
\section{SceneCollisionNet Evaluation}
We benchmark \algname against 4 baseline point cloud collision algorithms using synthetic data on two tasks: (1) a dataset of 1000 scene/object pairs with 2048 queries per scene/object pair where objects move on 16 linear trajectories through a scene, and (2) a dataset of 192,000 total grasps using the Franka Panda gripper~\cite{eppner2021icra}, gathered from 5 scenes and four object categories (mugs, cylinders, boxes, and bowls), where each scene has between 7 and 10 objects from the same category. In both tasks, ground truth collisions between the robot and the scene are calculated using FCL and the mesh models of the objects and robot in the scene. For each method and task, we compare the overall prediction accuracy and the computation time per query or grasp. We additionally report average precision (AP) scores, a weighted mean of precisions achieved at each recall threshold, for the trajectory benchmark and precision and recall for the grasp benchmark.

\subsection{Baseline Algorithms}
We benchmark \algname against both analytic and learned baselines. Marching Cubes baseline methods first create a mesh representation of the scene, object, or both from the raw point clouds~\cite{lorensen1987marching}. Signed distance function (SDF) methods use the Kaolin library~\cite{kaolin2019arxiv} for mesh to SDF conversion and GPU-based SDF evaluation. The baselines are:
\begin{enumerate}
    \item \textbf{Marching Cubes + SDF Scene (MC+SDFS)}: The points belonging to the object point cloud are transformed and evaluated using the scene SDF. If any point has distance, the object is in collision with the scene. 
    \item \textbf{Marching Cubes + SDF Object (MC+SDFO)}: The points belonging to the scene point cloud are transformed and evaluated using the object SDF. If any point has a zero or negative distance, the object is in collision with the scene.
    \item \textbf{Marching Cubes + FCL (MC+FCL)}: Collisions between the scene meshes and object meshes are determined using the flexible collision library (FCL)~\cite{pan2012fcl}. For a fairer comparison, we parallelize this method across 10 processes. We also show performance when it receives points directly sampled from the underlying object meshes (FO).
    \item \textbf{Pointnet Grid}: The scene is divided into coarse overlapping voxels; the object and each voxel are featurized using Pointnet++ set abstraction layers~\cite{qi2017pointnet++}, but no voxel convolution layers, and trained on the same dataset used for \algname. Predictions are averaged across the 8 corresponding voxels. This algorithm can be viewed as an ablation of \algname that does not share information between voxels via voxel convolution.
\end{enumerate}

\subsection{Results}
\algname outperforms all baselines in the linear trajectory collision environment, as shown in Table~\ref{tab:collisionbenchmark}, with $9.8\%$ and $15.8\%$ gains in accuracy and AP score, respectively, over the FCL baseline. Additionally, \algname is nearly 20 times faster than the parallelized FCL baseline, taking only about $10$ $\mu$s per collision query. \algname can predict over 500,000 queries in a single forward pass on an NVIDIA GeForce RTX 2080 Ti GPU, further reducing the time per query for large batches of queries.

The comparison with Pointnet Grid suggests the benefits of both the coarse voxel representation and the ability to share information between voxels via convolution. If the scene is encoded in the same way as the object (Pointnet++ layers only), the network fails to converge. When encoding coarse independent voxels using Pointnet++ set abstraction layers, but without using voxel convolutions to share information between them, the accuracy and AP scores are 16.5\% and 6.2\% lower, respectively.

Table~\ref{tab:graspbenchmark} shows the results on the grasping benchmark. In addition to evaluating each model on the grasp poses, we evaluate the models on pre-grasp poses that are offset along the approach axis by 5 cm. The baseline methods slightly outperform or show similar performance to \algname on the grasp dataset with no offset, but \algname outperforms baselines on the 5 cm offset with a 1.5\% improvement in accuracy as well as a 9.4\% improvement in precision. These results suggest \algname can struggle to predict collisions for geometries that are very close to being in or out of collision, but dramatically improves with increasing distance between objects and learns underlying structure beyond the points in the scene, while the other methods are unable to account for gaps in point cloud data.

\begin{table}[t!]
    \vspace{1.5mm}
	\centering
	\begin{tabu} to \linewidth {X[2.5l]X[c]X[c]X[2.5c]} \toprule
		\textbf{Algorithm} & \textbf{Accuracy} & \textbf{AP} & \textbf{Time / Query (ms)} \\\midrule
		MC+SDFO & $70.2\%$ & $0.651$ & $27 \pm 12$ \\
        MC+SDFS & $80.0\%$ & $0.781$ & $24 \pm 2$ \\
		MC+FCL (10x) & $75.4\%$ & $0.824$ & $0.49 \pm 0.06$ \\
		MC+FCL (10x, FO) & $83.4\%$ & $0.832$ & $0.74 \pm 0.13$ \\
		% \algname (fully observed) & $91.6\%$ & $0.925$ & $0.983$ & $11 \pm 4$ \\
		PointNet Grid & $76.7\%$ & $0.928$ & $0.026 \pm 0.035$ \\
		\algname & $\bm{93.2\%}$ & $\bm{0.990}$ & $\bm{0.010 \pm 0.002}$ \\
		\bottomrule
	\end{tabu}
	\caption{Benchmark results for 1000 scene/object pairs, with 16 linear trajectories and 2048 queries for each pair (2,048,000 total queries). \algname outperforms parallelized baselines that reconstruct meshes even from fully observed point clouds and a learned ablation that does not share information between voxels.}
	\label{tab:collisionbenchmark}
% 	\vspace{-8mm}
\end{table}

\begin{table}[t!]
	\centering
	\begin{tabu} to \linewidth {X[1.75l]X[c]X[c]X[c]X[1.25c]} \toprule
		\textbf{Algorithm} & \textbf{Accuracy} & \textbf{Precision} & \textbf{Recall} & \textbf{Time (ms)} \\\midrule
		MC+SDFO & $90.8 / 81.2$ & $31.1 / 59.2$ & $\bm{95.4} / \bm{98.9}$ & $62$\\
        MC+SDFS & $\bm{94.4} / 78.2$ & $\bm{32.0} / 63.2$ & $12.0 / 58.6$ & $37$ \\
		MC+FCL (10x) & $\bm{94.4} / 80.4$ & $27.8 / 63.6$ & $10.8 / 68.8$ & $0.27$ \\
		\algname & $92.4 / \bm{82.7}$ & $21.2 / \bm{73.0}$ & $19.3 / 71.8$ & $\bm{0.018}$\\
		\bottomrule
	\end{tabu}
	\caption{Benchmark results for 192,000 grasps across 20 scenes of 4 object categories and offsets of 0 cm / 5cm. The baselines slightly outperform \algname for the 0 cm offset, but \algname outperforms baselines in both accuracy and precision for the 5 cm offset while recalling over 70\% of the collision-free grasps 15x faster.}
	\label{tab:graspbenchmark}
% 	\vspace{-8mm}
\end{table}

%% file: includes/7-policy-results.tex
\section{Policy Evaluation}
We evaluate the proposed MPPI policy in both simulation and in physical tabletop scenes, recording the number of successful grasps and placements in each scenario as well as the time taken for picking and placing each object. We use $T=300$, $H=40$, and $\Sigma = 0.3\cdot I$ and query the policy at 1 Hz. 

\subsection{Simulation Evaluation}
We compare \algname to MC + FCL (10x), the best performing baseline, as part of the MPPI policy in 10 simulated scenes with 10 objects each, drawn from a dataset of bowls, mugs, cylinders, and boxes. The objects are arranged randomly in a stable pose and not in collision, but may not be graspable. An object order is selected randomly, and the policy is given grasps on the specified target object to grasp and place that object in a different location on the table. The policy is given two attempts for each object, and if it is unable to pick or place the object, it moves on to the next target. In total, the policies interact with the scenes for 4.5 hours. Results in Table~\ref{tab:simresults} suggest that \algname can dramatically speed up the MPPI policy, which can rearrange over half of the objects.

\begin{table}[t!]
    \vspace{1.5mm}
% 	\centering
	\begin{tabu} to \linewidth {X[3l]X[c]X[c]X[3c]} \toprule
		\textbf{Algorithm} & \textbf{Grasps} & \textbf{Placements} & \textbf{Time (min)} \\\midrule
		MC+FCL (10x) & $109$ & $92$ & $164$ \\
		\algname & $\bm{110}$ & $\bm{99}$ & $\bm{100}$ \\
		\bottomrule
	\end{tabu}
	\caption{Simulation rearrangement results when using \algname and MC+FCL (10x) as collision checkers in the MPPI policy. \algname speeds up scene interaction and leads to more placements.}
	\label{tab:simresults}
% 	\vspace{-3.5mm}
\end{table}

\subsection{Physical Evaluation}
We additionally evaluate the MPPI policy with \algname on a set of 10 physical tabletop scenes with a Franka Panda robot and an Intel RealSense LiDAR Camera L515. We divide the scenes into two categories: barrier scenes and rearrangement scenes, each with between 3 to 11 YCB objects. Examples are shown in Figures~\ref{fig:rearrangement},~\ref{fig:placements} and~\ref{fig:rollouts}. In barrier scenes, a tall box divides the scene and objects must iteratively be grasped and placed on the opposite side of the barrier. Rearrangement scenes are similar to the simulated scenes, where objects are placed randomly on the table. However, in this case, placement is also restricted to a single side of the scene and both grasps and placements must be made among clutter. Placement becomes more difficult later in trials when the zone fills with objects.

In the four barrier scenes, the policy grasps $16/17$ objects and places $12/17$ objects successfully. Three failures were due to collisions in the trajectory or incorrect placement choice, one was due to object motion in the gripper, and one was due to the policy being unable to find a placement. In the rearrangement scenes, $25/27$ grasps and $20/27$ placements were successful. Of these failures, five were due to collision errors in the trajectory or placement position, with one failure each due to motion in the hand and no placements found. 

The policy's grasping performance suggests it can consistently generate collision-free robot trajectories to specified goals in the presence of both clutter and a challenging divider that requires planning to significantly deviate from a straight-line trajectory. The placement performance indicates that the addition of checking collisions with an object in the hand makes finding collision-free trajectories more difficult, but the policy is still able to effectively reason about collisions along the trajectory and at the placement location.

%% file: includes/8-conclusion.tex
\section{Conclusion}
We present a learned collision checking model that dramatically increases collision checking speeds between point clouds for motion planning in real-world rearrangement tasks. While we focus our evaluation on static scenes in this paper, the supplementary video also shows a dynamic example where an obstacle is encountered during placement. While the MPPI and \algname framework can support a higher control frequency than 1 Hz, the policy reaction time is limited by the segmentation and point cloud processing; in future work, we will investigate the ability of the policy to react to more dynamic environments, other ways of generating candidate trajectories, and adaptation to constrained environments such as shelves and cabinets.